\documentclass[final,5p,times]{elsarticle}

\journal{Medical Image Analysis}
\biboptions{authoryear,round}

\let\cite\citep

\usepackage{mathnotation}
\usepackage{dblfloatfix} 
\usepackage{amstext}
\usepackage{amsmath}
\usepackage{amssymb}
\usepackage{bm}
\usepackage[english]{babel}
\usepackage{graphicx}
\usepackage{xurl} 
\usepackage{soul} 
\usepackage{xcolor}

\usepackage{array}
\usepackage{booktabs}

\usepackage{physics}
\usepackage{tikz}
\usepackage{mathdots}
\usepackage{multirow}
\usepackage{algorithm}
\usepackage{algpseudocode}
\usepackage{caption}
\usepackage{float} 
\usepackage{microtype}
\usepackage{placeins}

\graphicspath{{Figures/}}

\begin{document}

\begin{frontmatter}

\title{IRSDE-Despeckle: A Physics-Grounded Diffusion Model for Generalizable  Ultrasound Despeckling}

\author[inst1]{Shuoqi Chen}
\author[inst1]{Yujia Wu}
\author[inst1]{Geoffrey P. Luke\corref{cor1}}

\address[inst1]{Thayer School of Engineering, Dartmouth College, Hanover, NH, USA}

\cortext[cor1]{Corresponding author}
\ead{geoffrey.p.luke@dartmouth.edu}

\begin{abstract}

Ultrasound imaging is a cornerstone in medical diagnostics due to its real-time capabilities and non-invasive nature. However, the inherent presence of speckle noise and other artifacts can significantly degrade image quality, complicating image interpretibility and diagnosis. Effective despeckling techniques are thus essential to enhance image clarity and reliability. Recent advancements in generative models, particularly diffusion models, show promise in image restoration tasks. Diffusion models operate by modeling the data distribution through a gradual noising and denoising process, effectively capturing complex data structures. Building upon the Image Restoration Stochastic Differential Equations framework, our approach introduces a novel despeckling method for ultrasound images. We curated large paired datasets by generating simulated ultrasound images from speckle-free magnetic resonance imaging scans using the Matlab UltraSound Toolbox, which models ultrasound physics to produce realistic signal generation. Our model was trained to reconstruct speckle-free images at the original spatial resolution, while preserving anatomically meaningful edges and contrast. Evaluated on a separately simulated ultrasound test set, our method consistently outperforms both classical filtering approaches and recent learning-based despeckling methods. We quantified the model prediction uncertainty using cross-model variance and showed that higher uncertainty aligns with higher reconstruction error, providing a practical indicator of difficult or failure-prone regions. We also evaluated sensitivity to simulation settings and observed domain shift, highlighting the need for diversified training or adaptation to improve robustness and reliability for clinical deployment. Overall, our method combines physics grounded training data, diffusion based image restoration, and uncertainty analysis into a practical pipeline that reduces ultrasound speckle artifacts and improves interpretability for robust clinical decision making. 
 
\end{abstract}


\end{frontmatter}

\begin{figure*}[t]
\centering
\includegraphics[width=0.95\textwidth]{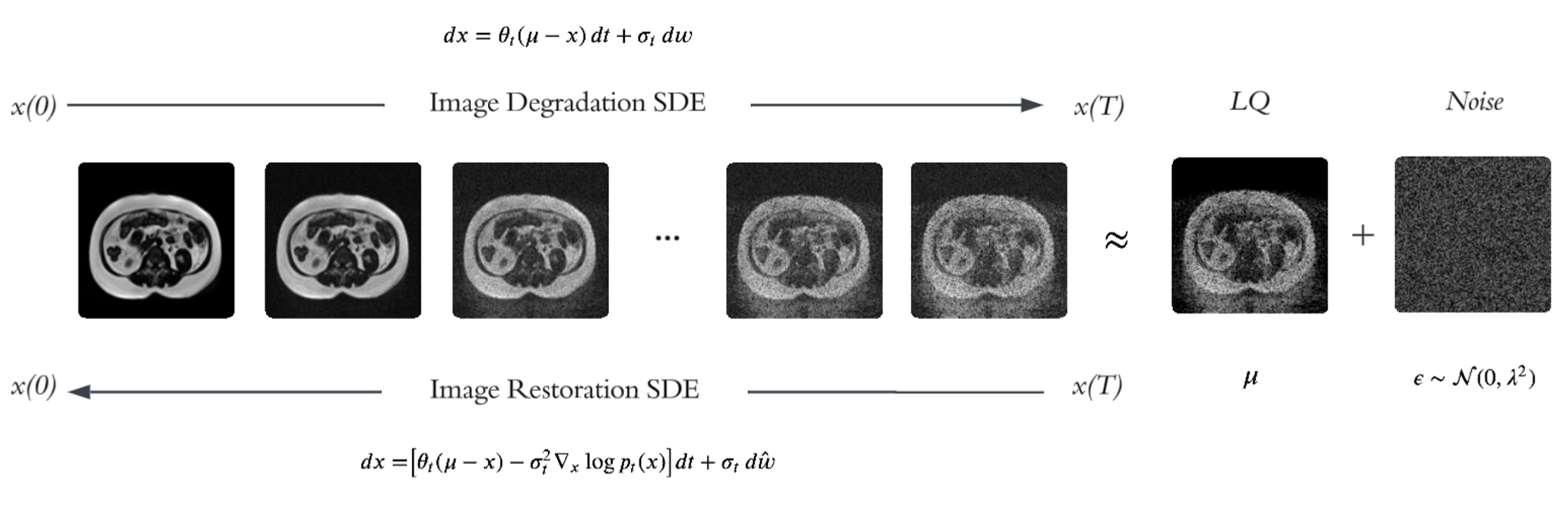}

\caption{Overview of our Image Restoration SDE (IRSDE) ultrasound despeckling pipeline. The forward {image degradation} SDE progressively perturbs a HQ image $x(0)$ into a terminal state $x(T)$, which is approximately a noisy observation of the low-quality image. The reverse {image restoration} SDE then evolves $x(T)$ back towards $x(0)$ by simulating the corresponding reverse-time dynamics, where the drift includes the score term $\nabla_x \log p_t(x)$, which is parameterized via a conditional time dependent U-Net noise predictor $\hat{\epsilon}_\phi(x_t,\mu,t)$.}
\label{fig:despeckling_overiew}
\end{figure*}


\section{INTRODUCTION}


Despeckling in ultrasound imaging addresses the challenge of speckle noise. Unlike additive Gaussian noise or sensor noise, speckle is multiplicative and structure-dependent in nature; it arises from coherent interference of echoes from many sub-resolution scatterers within a resolution cell \cite{49_Burckhardt1978,50_Wagner1983}. 
The random spatial distribution of these scatterers causes their backscattered waves to interfere constructively and destructively, producing the characteristic mottled appearance of ultrasound images \cite{49_Burckhardt1978,48_Narayanan2009}.

Classical approaches to ultrasound despeckling image include processing techniques such as anisotropic diffusion, non-local means, and wavelet-based filtering \cite{34_unknown,48_Narayanan2009,MichailovichTannenbaum2006Despeckling}. These methods aim to suppress speckle while preserving anatomical boundaries, for example by restricting diffusion across edges or selectively filtering frequency bands in the wavelet domain \cite{34_unknown,MichailovichTannenbaum2006Despeckling}. While effective in certain cases, these methods often oversmooth fine textures, introduce artifacts, and require careful parameter tuning, limiting their practical use\cite{48_Narayanan2009}. 

To overcome the limitations of classical despeckling filters, learning-based methods for ultrasound despeckling have increasingly adopted data-driven approaches. The U-Net architecture became a strong baseline due to its encoder-decoder design with skip connections that preserve fine structural details \cite{Dong2021FDCNN,Cammarasana2022RealtimeUSDenoising}. Variants such as residual dense U-Nets improved performance by introducing residual and dense connections, which ease training of deeper networks and improve texture reconstruction \cite{gan_rw_zhang2022}. Attention mechanisms, including spatial and channel attention, have also been integrated to adaptively emphasize informative features and suppress speckle in homogeneous regions. These modifications consistently reported better preservation of organ boundaries and subtle structures compared to plain U-Net models \cite{LanZhang2020MARU,Ardakani2024AdaRes}.

Generative Adversarial Networks (GANs) further reframe ultrasound despeckling as a conditional image-to-image translation problem, in which a generator synthesizes candidate despeckled images while a discriminator enforces realism and anatomical plausibility \cite{45_unknown,31_unknown}. Representative approaches employ U-Net–based generators with residual or dense connections and joint adversarial–reconstruction losses, demonstrating improved speckle suppression and perceptual quality. Other variants leverage cross-modality translation, for example using pix2pix-style frameworks, to map ultrasound images to magnetic resonance imaging (MRI) targets to approximate a cleaner reference domain while preserving anatomical structure \cite{Vieira2024GANTransfer}. Despite producing visually appealing results, GAN-based methods suffer from fundamental limitations: adversarial training can be unstable and prone to mode collapse, and hallucinated or distorted structures may compromise diagnostic reliability \cite{Sikhakhane2024GANReviewUSDenoising}.

Diffusion models offer a paradigm shift from GANs by modeling despeckling as an iterative denoising process rather than a single feed-forward pass \cite{Asgariandehkordi2023IUSDDPM,Asgariandehkordi2024TUFFCDDPM,GuhaActon2023SDDPM}. A forward process gradually corrupts the images with additive white Gaussian noise, and a learned reverse process progressively removes the noise while conditioning on timestep embeddings \cite{Ho2020DDPM,Song2020Score}. Early applications have demonstrated that diffusion-based methods can preserve characteristic speckle statistics while improving image clarity beyond classical filters \cite{GuhaActon2023SDDPM,Asgariandehkordi2023IUSDDPM,Asgariandehkordi2024TUFFCDDPM,dehazing_ultrasound_diffusion}. Although diffusion models mitigate GAN failure modes such as mode collapse by learning the data distribution through an iterative process, they are not immune to hallucination, especially when training data are insufficiently representative \cite{dehazing_ultrasound_diffusion}. Incorporating physically grounded training data and strong data-consistency constraints is therefore critical for diagnostic reliability \cite{41_unknown}.

Taken together, existing approaches highlight clear progress but also expose a gap between strong empirical performance and physically consistent, data-efficient learning. To close this gap, we introduce a physics-informed diffusion framework for ultrasound despeckling. To address the lack of supervised training data, we first generated a high-quality synthetic dataset using the Matlab UltraSound Toolbox (MUST), where MRI slices serve as clean anatomical references and were simulated into realistic B-mode images with physically grounded speckle. Building on this dataset, we adapted the Imaging Restoration Stochastic Differential Equation (IRSDE) framework to model the degradation from clean to speckled images as a mean-reverting stochastic process, and trained the corresponding reverse SDE to recover MRI-like speckle-free images \cite{Luo2023IRSDE}. This enabled an ultrasound despeckling model that is physically consistent and capable of producing either a single high-quality estimate or multiple estimates for ensembling. We evaluated our model on both an out-of-distribution test dataset and publicly available US images. Results show that our method outperforms classical and learning-based approaches in peak signal-to-noise ratio (PSNR), structural similarity (SSIM), and language perceptual image patch similarity (LPIPS) metrics, all while offering improved boundary preservation and reduced speckle appearances without the loss of subtle structure.
\section{Methods}

\subsection{SDE-based Image Restoration Diffusion Modeling}


Building on the original discrete Denoising Diffusion Probabilistic Models (DDPM) formulation, stochastic differential equations extend diffusion modeling to continuous time, providing a more flexible sampling strategy.
The SDE formulation generalizes the original discrete models and enables more flexible sampling strategies, such as using arbitrary step sizes, advanced numerical solvers, or even deterministic probability-flow ODEs, while preserving the underlying generative behavior \cite{Ho2020DDPM,Song2020Score}. By characterizing the continuous evolution of a random variable x(t), the SDE formulation thereby offers finer control over the denoising dynamics and improves capacity for generalization across imaging modalities. 

The IRSDE model further builds on the continuous-time diffusion framework by formulating image restoration as a reverse-time stochastic process \cite{Luo2023IRSDE}. Unlike conventional diffusion models that corrupt clean images using a predetermined Gaussian noise schedule, IRSDE introduces a mean-reverting forward SDE that explicitly drives a high-quality (HQ) image toward the observed low-quality (LQ) measurement. This construction produces a degradation path that is physically meaningful and tightly coupled to the actual observation, which in turn leads to more data-consistent forward simulations and more stable score estimation during training.

In IRSDE, the forward process is designed to transform a clean HQ image $\mathbf{x}_0$ to a degraded LQ counterpart $\mu$ using a drift term that pulls the sample toward $\mu$, combined with a diffusion term that adds Gaussian uncertainty. The forward SDE is defined as: 

\begin{equation}
    d\mathbf{x}_t = \theta_t(\mu - \mathbf{x}_t)dt + \sigma_t d\mathbf{w}_t,
\end{equation}

\noindent where $\mathbf{x}_t$ denotes the image state at diffusion time t, $\theta_t$ is a time-dependent drift coefficient, $\sigma_t$ is the noise amplitude, and $d\mathbf{w}_t$ is an increment of a standard Wiener process. It is important to note that the Gaussian perturbations here are not a way to inject speckle noise, but rather a mathematically tractable mechanism to define the forward and reverse dynamics. By progressively corrupting $\mathbf{x_0}$ towards the LQ reference $\mu$, the model learns to map noisy observations back to clean ones. The real speckle information is presented at the low quality reference LQ. In practice, the Gaussian corruption serves as a flexible training surrogate \cite{Song2020Score} that enables stable likelihood-based optimization, while the conditioning on $\mu$ ensures that the learned denoising process aligns with the structured, non-Gaussian characteristics of the HQ image. This ensures that the trajectory remains anchored to the degraded observation rather than just purely random. {Over time, the solution converges to a terminal state $\mathbf{x_{T}}$ that follows a Gaussian distribution centered at $\mu$:}

\begin{equation}
    \mathbf{x}_T \sim \mathcal{N}(\mu, \lambda^2),
\end{equation}

\noindent {where $\lambda^2$ reflects the cumulative diffusion strength.}

This statistical approximation enables a tractable formulation of the reverse-time SDE. The reverse SDE used to reconstruct the HQ image from $\mu$ is given by:

\begin{equation}
    d\mathbf{x}_t = [\theta_t(\mu - \mathbf{x}_t) - \sigma_t^2 \nabla_{\mathbf{x}_t} \log p_t(\mathbf{x}_t)] dt + \sigma_t d\bar{\mathbf{w}}_t,
\end{equation}

\noindent where the score function $\nabla_{\mathbf{x}_t} \log p_t(\mathbf{x}_t)$ represents the gradient of the log-density, and $d\bar{\mathbf{w}}_t$ is a reverse-time Wiener process.


To improve restoration stability and fidelity, IRSDE incorporates a maximum likelihood objective that directly aligns the reverse process with high-quality target images. Overall, the framework leverages continuous-time iterative refinement, explicit conditioning on the degraded observation, and implicit score-function estimation to achieve strong noise suppression while preserving details.

Following the success of IRSDE in natural image restoration, we extend the framework to ultrasound despeckling, with an overview of the proposed pipeline provided in Figure~\ref{fig:despeckling_overiew}. Prior work shows that IRSDE generalizes well across tasks such as deraining, deblurring, and denoising, demonstrating robustness to diverse and complex noise structures \cite{Luo2023IRSDE}. This adaptability is particularly valuable for ultrasound, where speckle in B-mode ultrasound follows a non-normal distribution, is signal-dependent, and is tightly coupled to underlying tissue structure. By conditioning the forward process to the speckled image, IRSDE provides a physically meaningful framework to suppress speckle pattern while preserving fine anatomical details.



\subsection{High-quality Target Data Generation}

\begin{figure*}[t]
\centering
\includegraphics[width=0.99\textwidth]{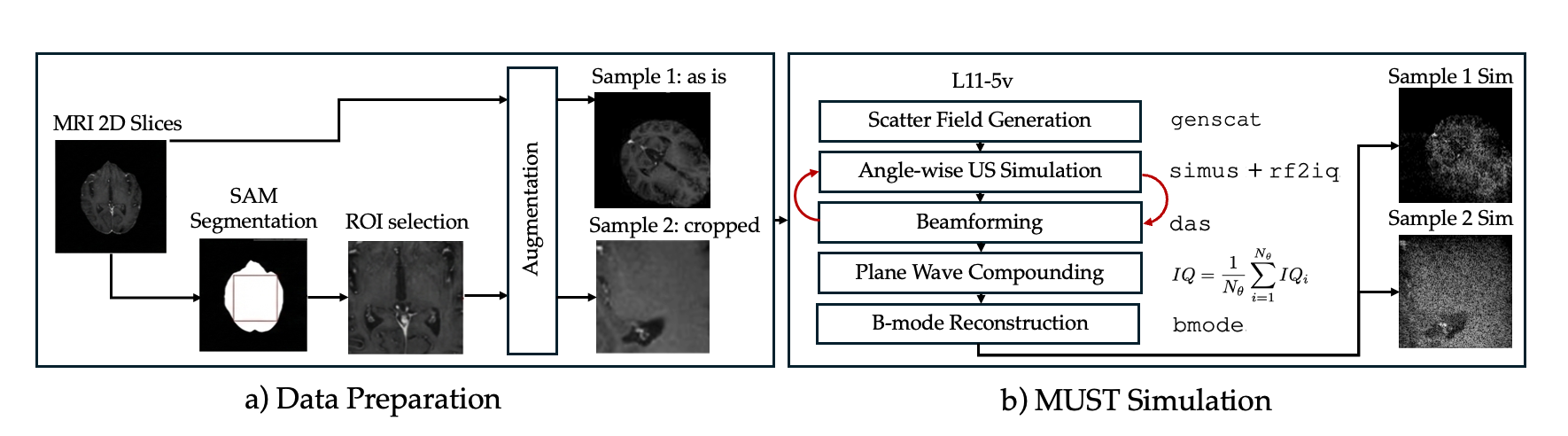}
\caption{Overview of our data generation scheme. a) MRI slices were segmented and augmented to maximize anatomically informative content. b) The processed 2D grayscale MRI samples were simulated into B-mode US images using the MUST toolbox. The pipeline output paired high quality and low quality images for training.}
\label{fig:data_processing}
\end{figure*}

To overcome the absence of true speckle-free ultrasound image targets, we utilized a physics-driven data generation pipeline to obtain paired HQ--LQ examples. We simulated ultrasound images using MRI slices as the clean HQ targets, as they are inherently speckle-free. Figure.~\ref{fig:data_processing} shows our holistic data generation pipeline. We used several publicly available MRI datasets, including MRNetKneeMRIs \cite{MRNet}, Duke Liver Dataset v2 \cite{DukeLiverV2}, BrainMetShare-3 \cite{BrainMetShare}, and Duke-Breast-Cancer-MRI \cite{DukeBreastCancerMRI} to build our training set in order to enhance anatomic generality. To ensure consistency across different sources, we extracted 2D slices along standard anatomical planes and converted all images to grayscale. 
 
Using full 2D MRI slices often resulted in images with large black background regions surrounding the tissue, which were not representative of real clinical ultrasound images, and could introduce domain-shift artifacts during training. In addition, global anatomical patterns, such as overall organ shape, tended to be consistent across patients and imaging protocols due to shared human physiology. Reliance on these global cues risked overfitting, as the model might exploit superficial anatomical regularities rather than features relevant to image restoration.

To mitigate this, for each extracted MRI 2D slice, we additionally generate localized patches as a regularization mechanism to reduce dependence on global structural cues. In order to ensure that extracted patches contained meaningful anatomical information, we segmented regions of interest (ROIs) using the SAM2 model, a zero-shot vision transformer capable of producing accurate foreground masks without manual annotations \cite{SAM2}. Small spurious regions are removed using connected component analysis, retaining only the largest cohesive structure. Only the largest square bounding box which is fully contained within the foreground region was used, standardizing patch geometry while maximizing anatomical content and minimizing background. Both the full slices and their localized patches provide complementary benefits. Full slices preserve structure and anatomical context, while zoomed-in patches discourage the model from exploiting similar global patterns across samples. We thus include both full slices and extracted patches as HQ targets in the training set to increase data diversity.

Unlike typical machine learning pipelines where data augmentation can be applied on the fly during training, ultrasound simulation is computationally expensive and has to be performed offline. Consequently, we applied augmentation to the HQ target before simulating its LQ counterpart. This augmentation is important because using unaugmented HQ inputs for ultrasound simulation would yield scatterer patterns with similar scale and spatial structure across samples containing similar looking anatomy, leading to unintended structural information leakage. Since geometric transformations do not affect the underlying physics of ultrasound scattering, we applied standard augmentations, including flips, rotations, and rescaling, to increase anatomical diversity prior to simulation. We intentionally avoided augmentations that degraded image quality (e.g., blurring, noise injection, brightness shifts), as the HQ images serve as the restoration targets and have to remain artifact-free.

\subsection{Low-quality Ultrasound Data Simulation}

After constructing and augmenting the HQ targets, we then simulated their corresponding LQ counterparts using a physics based plane wave compounding pipeline. The simulation was implemented in MATLAB using the MUST toolbox, which provided numerical models for ultrasound wave propagation, scatterer generation, beamforming, and B-mode image formation \cite{MUST}. An \texttt{L11-5v} linear array transducer model was used for the simulations. To construct the scattering medium, we loaded pre-generated HQ images and interpreted their intensities as a spatial map of scatterer intensity. Scatterers were generated by sampling spatial locations $(x_s, z_s)$, with reflection coefficients drawn from Rayleigh distributions whose local means were modulated by the MRI grayscale values. Redundant or spatially overlapping scatterers were pruned to reduce computational cost. 

The formation of ultrasound images was performed by transmitting plane waves at 15 steering angles uniformly sampled from $\left[-\frac{7\pi}{32},\, \frac{7\pi}{32}\right]$. For each angle, the acoustic wave propagation was numerically simulated, and raw radio-frequency (RF) channel data were generated. The RF signals were then demodulated into in-phase and quadrature (IQ) components, followed by delay-and-sum beamforming. The resulting IQ images were compounded across all steering angles, which improved contrast resolution and reduced speckle. After compounding, envelope detection and logarithmic compression with a 40-dB dynamic range were applied to generate the final simulated B-mode ultrasound images at a resolution of $128 \times 128$ pixels.

\subsection{Training Pipeline}

\begin{algorithm}[t]
\caption{IRSDE-Despeckle Training with Normalization}
\label{algo:irsde-training}
\begin{algorithmic}[1]
\Require Paired data $(x_0, \mu)$; total steps $T$; schedules $\{\theta_t, \sigma_t\}_{t=1}^T$ with $\sigma_t^2/\theta_t = 2\lambda^2$; network $\hat{\epsilon}_\phi(x_t,\mu,t)$
\Repeat
  \State Sample a minibatch $\{(x_0^{(b)}, \mu^{(b)})\}_{b=1}^B$
  \State Normalize inputs
  \State Sample $t \sim \mathrm{Uniform}\{1,\dots,T\}$
  \State Compute integrated drift $\bar{\theta}_t = \int_0^{t}\theta_z\,dz$
  \State Compute mean and variance:
    \[
      \tilde{m}_t = \tilde{\mu} + (\tilde{x}_0 - \tilde{\mu})\,e^{-\bar{\theta}_t}, \qquad
      \tilde{v}_t = \lambda^2\bigl(1 - e^{-2\bar{\theta}_t}\bigr)
    \]
  \State Sample $\epsilon \sim \mathcal{N}(0,I)$ and set
    \[
      \tilde{x}_t = \tilde{m}_t + \sqrt{\tilde{v}_t}\,\epsilon
    \]
  \State Predict noise $\hat{\epsilon} = \hat{\epsilon}_\phi(\tilde{x}_t,\tilde{\mu},t)$
  \State Update $\phi$ by gradient descent:
    \[
      \mathcal{L}(\phi) = 
      \mathbb{E}_{t,(x_0,\mu),\epsilon}\bigl[
      \|\epsilon - \hat{\epsilon}_\phi(\tilde{x}_t,\tilde{\mu},t)\|_2^2
      \bigr]
    \]
\Until{converged}
\end{algorithmic}
\end{algorithm}

Our training pipeline and network architecture largely followed the IRSDE formulation \cite{Luo2023IRSDE}, adapted to single-channel grayscale ultrasound inputs. In brief, a forward SDE progressively perturbed the clean target toward its degraded counterpart, and a time-conditioned score network was trained to predict the injected noise at arbitrary timesteps. Building on this principle, we trained on paired grayscale images $x_0,\mu\in\mathbb{R}^{1\times H\times W}$. At each iteration, we sampled a timestep $t\sim\mathrm{Uniform}\{1,\ldots,T\}$ with $T=100$. We then sampled $\epsilon\sim\mathcal{N}(0,I)$, where $I$ is the identity matrix, and used the IRSDE forward process to generate the perturbed state $x_t$. The denoiser took the concatenated input $[x_t\mid\mu]\in\mathbb{R}^{2\times H\times W}$ and predicted the injected noise $\hat{\epsilon}_\phi(x_t,\mu,t)\in\mathbb{R}^{1\times H\times W}$. Model parameters were optimized by minimizing the noise-prediction loss computed as the mean absolute error over pixels:

\begin{equation}
\mathcal{L}(\phi)=\mathbb{E}\bigl[\|\epsilon-\hat{\epsilon}_\phi(x_t,\mu,t)\|_1\bigr],
\label{eq:noise_pred_loss}
\end{equation}

Before applying the forward perturbation, we normalized both $\mu$ and $x_0$ to zero mean and unit variance using per-sample statistics, as summarized in Algorithm~\ref{algo:irsde-training}. We found this normalization stabilized optimization and accelerated convergence by improving the conditioning of the learning problem and enforcing scale invariance. This encouraged the network to learn structural rather than intensity-dependent representations and to become less sensitive to arbitrary grayscale scaling. Moreover, normalization aligned the data distribution with the assumptions of diffusion-based formulations, in which the noise schedule operated on normalized units.

The noise-prediction network was implemented as a time-conditioned, four-resolution-level U-Net with skip connections. Starting from a base of 64 channels, the feature dimension increased by factors of 1, 2, 4, and 8 across encoder stages, reaching 512 channels at the deepest level, and then mirrored symmetrically in the decoder. Each resolution stage comprised residual blocks with Swish activations and GroupNorm of 32 groups, with the timestep embedding added within every block to condition feature responses on the noise level. BatchNorm was not used, and a self-attention module was included at the $16\times16$ bottleneck to better capture long-range dependencies. The timestep embedding itself was computed using a sinusoidal positional encoding followed by a lightweight fully connected network.

We optimized the network using Adam with learning rate $2\times10^{-4}$, $(\beta_1,\beta_2)=(0.9,0.999)$, zero weight decay, and batch size 4. Training was performed on approximately $8{,}000$ paired LQ-HQ patches of size $128\times128$. Following the IRSDE training protocol, we trained for $400{,}000$ iterations and applied a learning-rate decay schedule during optimization, with initial learning rate of $10^{-4}$. We evaluated the trained model on the held-out Uterine Myoma MRI dataset (UMD), which was entirely unseen during training \cite{UMD2024}. As in the training setup, LQ--HQ pairs for evaluation were generated using the MUST simulator with the same parameter settings.


\subsection{Uncertainty Estimation via Cross-validation Ensemble}
\label{subsec:uncertainty_method}

To evaluate epistemic uncertainties of our model prediction, we trained our method using a 5-fold cross-validation ensemble and calculated the cross-model variance. This 5-fold ensemble served two complementary purposes. First, training on multiple data partitions and evaluating across held-out folds provided a more reliable estimate of model generalization under dataset variability. The independently trained fold models formed a cohort whose prediction diversity reflected uncertainty arising from limited data support. Second, aggregating predictions from independently trained fold models allowed us to construct an ensemble whose averaged output reduced variance and counteracted individual model biases. Ensembles often provide strong estimates of epistemic uncertainty by combining predictions from multiple independently trained models that converge to distinct solutions in parameter space. Prior work has similarly found that ensembles achieved reliable aleatoric and epistemic uncertainty disentanglement \cite{ValdenegroToro2022DeeperLook}.

To assess whether the epistemic uncertainty predicted the reconstruction error, we analyzed the relationship between per-image predictive variance and per-image mean squared error (MSE) with respect to the HQ target. For each image, we computed the per-image MSE by averaging pixel-wise squared error. We then quantified linear association using Pearson's correlation coefficient, which was invariant to affine rescaling and directly measured the linear association between two continuous quantities. In our experiments on the reserved 400-image test set generated from the UMD, we reported per-image correlation rather than per-pixel correlation because large homogeneous background regions could otherwise dominate the statistics.



\section{Results}
\label{sec:results_discussion}

\begin{figure*}[t]
\centering
\includegraphics[width=0.95\textwidth]{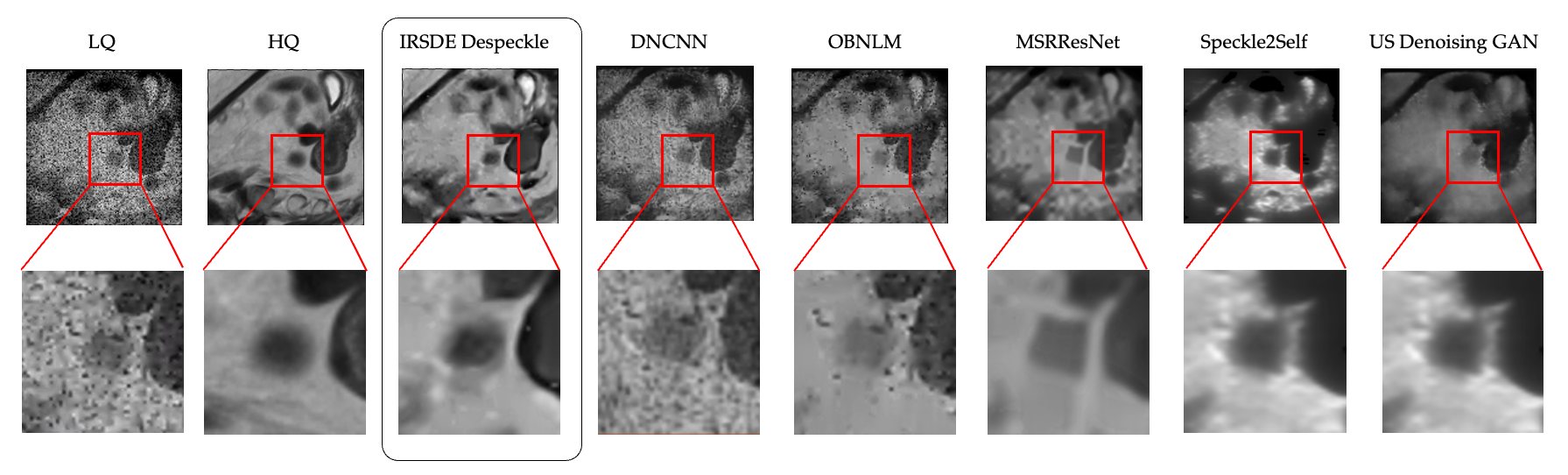}
\caption{Visual comparison of despeckling on a simulated ultrasound image from the held-out test dataset from UMD. Columns show the paired LQ--HQ images, our IRSDE despeckling result, and representative comparisons (DnCNN, OBNLM, MSRResNet, Speckle2Self, and ultrasound denoising GAN). The red boxes in the top row mark an example region of interest, and the corresponding zoomed-in croped images are shown in the bottom row. Our method suppresses speckle while preserving edges and contrast in the highlighted region, whereas several comparison SOTA methods either leave residual speckle or oversmooth fine structures and attenuate local contrast.}
\label{fig:close_up_view}
\end{figure*}

\subsection{Evaluation with Simulated Ground-truths}
\label{subsec:simulated_results}

We compared our method against a broad set of general-purpose denoisers and ultrasound-specific despeckling approaches, spanning classical filtering, learning-based natural-image restoration models, and recent generative methods designed for ultrasound despeckling. The quantitative comparison is summarized in Table~\ref{tab:method_categorization_full}, and a representative visual example is shown in Figure~\ref{fig:close_up_view}. 

Overall, IRSDE--Despeckle achieves the strongest performance across all metrics in Table~\ref{tab:method_categorization_full} on the held-out UMD simulated test set, which is generated from previously unseen biological samples and therefore contains tissue structures not observed during training. These quantitative advantages are also reflected qualitatively in Figure~\ref{fig:close_up_view}, where IRSDE–Despeckle suppresses speckle while preserving edges and local contrast in the highlighted region. Classical filtering methods, such as wavelet-thresholding variants and OBNLM~\cite{coupe2009_obnlm}, yield lower PSNR and SSIM, with higher LPIPS, compared with learning-based approaches. Among learning-based natural-image denoisers, DnCNN-3-Deblocking~\cite{DnCNN} provides the strongest baseline performance, while ultrasound-targeted baselines, including US Denoising GAN~\cite{gan_rw_zhang2022} and Speckle2Self~\cite{35_Li2025}, remain below IRSDE-Despeckle performance in all three metrics.

We also observed an additional gain from model ensembling, as shown in Table~\ref{tab:method_categorization_full}. The IRSDE–Despeckle ensemble was implemented via 5-fold cross validation on the training dataset, which yielded five models each trained on four folds with the remaining fold held out. At inference, we averaged the outputs of the five models, which further improved PSNR and SSIM and reduced LPIPS.

\begin{table*}[!t]
\centering
\caption{Quantitative comparison of different SOTA ultrasound despeckling methods in comparison to ours on an unseen simulated ultrasound test set from UMD}.
\label{tab:method_categorization_full}
\begin{tabular}{@{}lcccccc@{}}
\toprule
\textbf{Method} & \textbf{Learning-based} & \textbf{Natural Image} & \textbf{Ultrasound} & \textbf{PSNR$\uparrow$} & \textbf{SSIM$\uparrow$} & \textbf{LPIPS$\downarrow$} \\
\midrule
Wavelet Thresholding (Med)       &        & \checkmark & \checkmark & 15.73 & 0.44 & 0.43 \\
Wavelet Thresholding (Btw)       &        & \checkmark & \checkmark & 16.25 & 0.50 & 0.44 \\
Wavelet Thresholding (Ideal)     &        & \checkmark & \checkmark & 16.20 & 0.47 & 0.51 \\
OBNLM                            &        & \checkmark & \checkmark & 14.83 & 0.31 & 0.75 \\
MSResNet                         & \checkmark & \checkmark &  & 16.24 & 0.48 & 0.37 \\
DnCNN                            & \checkmark & \checkmark &        & 15.67 & 0.37 & 0.59 \\
DnCNN-3-Deblocking               & \checkmark & \checkmark &        & 16.33 & 0.48 & 0.30 \\
IMDN                             & \checkmark & \checkmark &        & 16.24 & 0.48 & 0.46 \\
IRCNN-Denoiser                   & \checkmark & \checkmark &        & 15.47 & 0.45 & 0.52 \\
SRMD                             & \checkmark & \checkmark &        & 16.25 & 0.48 & 0.35 \\
FFDNet                           & \checkmark & \checkmark &        & 14.78 & 0.31 & 0.77 \\
DPSR                             & \checkmark & \checkmark &        & 15.46 & 0.33 & 0.53 \\
Ultrasound Denoising GAN         & \checkmark &        & \checkmark & 12.44 & 0.23 & 0.59 \\
Speckle2Self                     & \checkmark &        & \checkmark & 15.68 & 0.44 & 0.31 \\
\midrule
\textbf{IRSDE Despeckle}          & \checkmark &        & \checkmark & \textbf{20.95} & \textbf{0.60} & \textbf{0.30} \\
\textbf{IRSDE Despeckle Ensemble} & \checkmark &        & \checkmark & \textbf{23.03} & \textbf{0.69} & \textbf{0.17} \\
\bottomrule
\end{tabular}
\end{table*}

\subsection{Evaluation with Real Ultrasound Images}
\label{sec:real_ultrasound_results}

\begin{figure}[!t]
\centering
\includegraphics[width=0.45\textwidth]{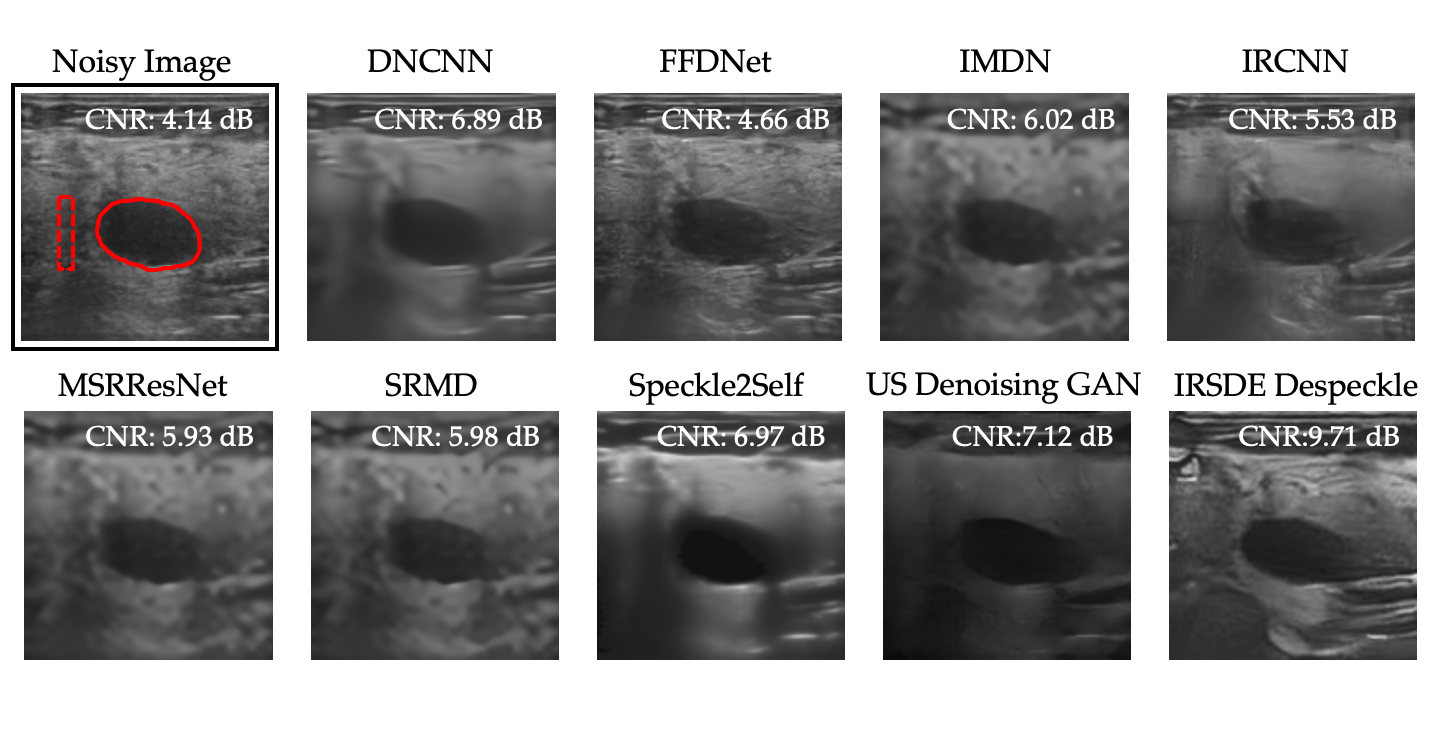}
\caption{Quantitative and visual comparison on a real ultrasound image from BrEaST Lesions USG dataset using contrast-to-noise ratio (CNR). For this visualization, CNR is shown in decibels as $\mathrm{CNR}_{\mathrm{dB}} = 20\log_{10}(\mathrm{CNR})$. The first panel shows the speckled input with the foreground ROI (solid red) and background ROI (dashed red) used for CNR computation. Remaining panels show the despeckled outputs from baseline methods and our method (IRSDE Despeckle), with the corresponding CNR value overlaid. Our method achieves higher CNR and better preserves lesion-to-background contrast while suppressing speckle, whereas several baselines either oversmooth and reduce contrast or leave residual texture that lowers CNR.}
\label{fig:referenceless_performance}
\end{figure}

We further evaluated our method on the publicly available BrEaST Lesions USG dataset, which contains real clinical ultrasound images acquired under varied clinical conditions and imaging settings \cite{breastlesions_usg}. Because the dataset does not provide despeckled ground truth, we compared methods qualitatively and complemented the visual assessment with no-reference quantitative measurements. Figure~\ref{fig:referenceless_performance} shows a representative example. Qualitatively, our method reduces speckle while maintaining clear lesion boundaries and preserving layered tissue structure. The lesion-to-background intensity separation remains well preserved, without excessive smoothing that would obscure fine anatomical detail.

In addition to these visual comparisons, we selectively computed no-reference metrics to quantify lesion contrast and background homogeneity. We computed contrast-to-noise ratio (CNR) using the annotated tumor mask as the foreground ROI, while forming the background ROI by aggregating locally homogeneous patches selected at a matched imaging depth, as illustrated in Figure~\ref{fig:referenceless_performance}. We also reported equivalent number of looks (ENL) and speckle signal-to-noise ratio (SNR) to characterize speckle suppression and background smoothness \cite{Baselice2017StatisticalSimilarityDespeckling,Wagner1983}, and included the blind/referenceless image spatial quality evaluator (BRISQUE) as a complementary no-reference perceptual metric \cite{Mittal2012BRISQUE}. Metric definitions, ROI selection details, and additional results are provided in Appendix Section~\ref{app:referenceless_metrics}.

\subsection{Prediction Uncertainties}

\label{subsec:uncertainty_results}

\begin{figure*}[t]
\centering
\includegraphics[width=0.9\textwidth]{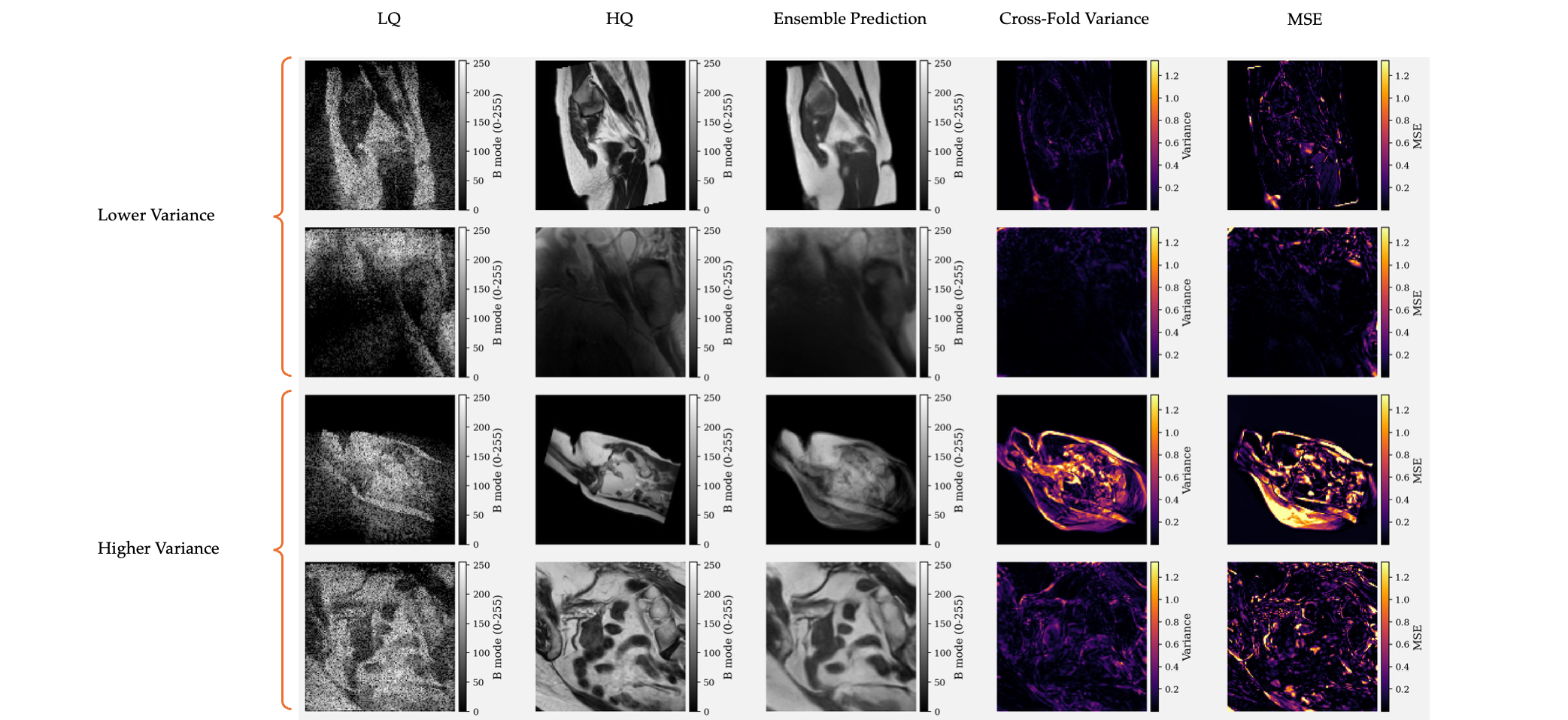}
\caption{Model prediction uncertainty and error visualization using a 5-fold cross-validation ensemble on selected samples from the MRNet knee MRI dataset. Examples with lower cross-fold variance are shown in the top rows, while examples with higher variance are shown in the bottom rows. Regions with higher variance indicate greater disagreement among fold models (higher epistemic uncertainty) and often coincide with localized hallucination artifacts and larger reconstruction errors. Images are displayed in 8-bit grayscale (0--255) for visualization, whereas MSE and variance are computed in the normalized image space.}
\label{fig:uncertainty}
\end{figure*}

Diffusion-based generative models are known to be susceptible to hallucinations. In the context of ultrasound despeckling, such hallucinations manifest as spurious edges, misplaced anatomical boundaries, or synthetic texture patterns. These artifacts emerge during the diffusion model’s iterative reconstruction, where each reverse step depends on the previous estimate. In regions that are ambiguous or weakly represented in training, small prediction errors at individual steps can accumulate and shift the trajectory toward visually plausible but anatomically incorrect structure. To probe whether such failures were accompanied by elevated epistemic uncertainty, we calculated cross-model variance using the 5-fold ensemble described in Section~\ref{subsec:uncertainty_method}.

Figure~\ref{fig:uncertainty} shows examples where the ensemble prediction exhibits noticeable differences from the HQ target. In regions where sharp boundaries in the ground truth transitions into more diffuse patterns in the simulated ultrasound, the model can produce non-existent, shadow-like structures. This effect is most pronounced in small, localized features where individual fold models disagreed and produced divergent predictions. Larger disagreement among ensemble models was accompanied by higher cross-model variance, which indicated elevated epistemic uncertainty.

Understanding this epistemic uncertainty provided a useful lens for interpreting model hallucination. In Figure~\ref{fig:uncertainty}, regions with both high reconstruction error and high variance across cross-fold ensembles reflected cases where the model ``knew it did not know''; these were typically out-of-distribution regions, rare anatomical configurations, or sharp structural transitions where training coverage was weak. The discrepancy between cross-model variance and reconstruction error was therefore a diagnostic of whether a failure arose from uncertainty or systematic misbehavior.

\begin{figure}[!t]
\centering
\includegraphics[width=0.5\textwidth]{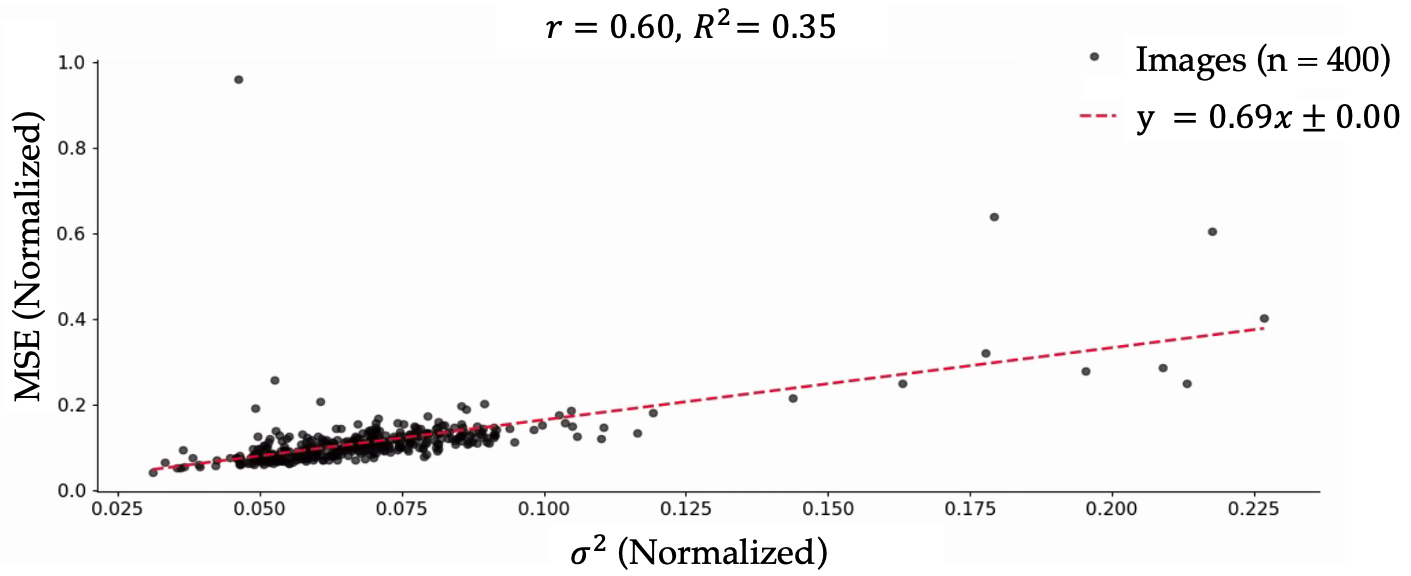}
\caption{Relationship between cross-model variance and reconstruction error (MSE) on the held-out UMD MRI test set. The x-axis represents the normalized per-image cross-fold predictive variance \(\sigma^{2}\) estimated from a 5-fold ensemble, while the y-axis represents the normalized per-image mean squared error with respect to the HQ target. The dashed line shows the least-squares linear fit with Pearson correlation \(r=0.60\), indicating that higher ensemble variance, and thus higher prediction uncertainty, is associated with larger reconstruction error.}
\label{fig:variance_mse_correlation}
\end{figure}

Figure~\ref{fig:variance_mse_correlation} illustrated the relationship between epistemic uncertainty represented by cross model variance and per image MSE on the held out UMD test set. The Pearson’s correlation was $ r= 0.60 $, which indicates a moderately strong positive linear relationship. This suggested that the variance between models increases in cases where the model performed worse, providing evidence that uncertainty was meaningfully aligned with error. Furthermore, the positive regression slope in Figure~\ref{fig:variance_mse_correlation} also showed that images with larger cross-fold variance tended to exhibit proportionally higher error, suggesting that the ensemble variance tracked despeckling difficulty.

\subsection{Sampling Step Reduction}

A central motivation of this work is to reduce inference latency toward practical, near real-time deployment. Diffusion-based despeckling restores the image through an iterative reverse trajectory, where speckle is progressively removed over many denoising steps. While this formulation is robust and yields high-quality reconstructions, the large number of iterations limits throughput in time-critical settings. In our implementation, sampling with $T=100$ steps requires an average of 1.1 seconds per image on a single NVIDIA RTX 4090 GPU, which remains a barrier to real-time clinical use. 

Table~\ref{tab:timesteps_results} summarizes the quantitative despeckling performance and runtime trade-off across sampling step counts; inference time decreases approximately linearly with the number of sampling steps. Moderate step reductions preserves most of the reconstruction quality; Reducing from $T=100$ to $T=50$ decreased inference time by half, with only a modest drop in PSNR and SSIM. In contrast, more aggressive reductions leads to rapid degradation in output quality, owing to the fact that a small number of sampling steps imposes a coarse numerical discretization of the reverse-time SDE. With too few steps, each update becomes large and noise errors compound.

\subsection{Generalization Across Simulated Probe Settings}
\label{subsec:cross_transducer_results}

On the held-out UMD test cases, we evaluated cross-transducer generalization by regenerating the simulated inputs in MUST under multiple probe configurations. In addition to the training probe \texttt{L11-5v}, we used \texttt{L12-3v} (linear), \texttt{C5-2v} (convex), and \texttt{P4-2v} (phased/sector), which span differences in operating frequency range and beam geometry and therefore induce shifts in speckle scale and spatial resolution.

\begin{figure}[!t]
\centering
\includegraphics[width=0.45\textwidth]{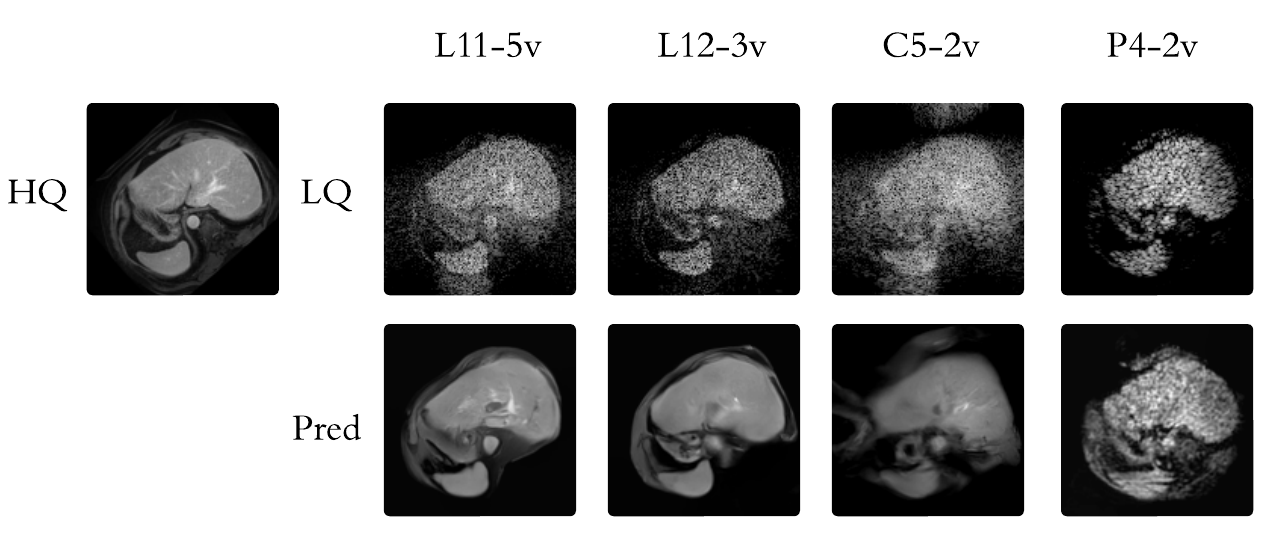}
\caption{Qualitative comparison of transducer specific degradation and restoration on the Duke Liver MRI dataset. The leftmost panel shows the high quality reference image, the top row presents simulated ultrasound images generated in MUST under different transducer settings (\texttt{L11-5v}, L12-3v, C5-2v, and P4-2v), and the bottom row shows the corresponding despeckled predictions produced by our method.}
\label{fig:transducer_settings}
\end{figure}

Our model, trained only on \texttt{L11-5v} simulated data, exhibits a clear cross-transducer domain shift at test time. Figure~\ref{fig:transducer_settings} illustrates qualitative differences in model performance across simulated ultrasound images from different probe settings. Predictions on ultrasound images simulated under the \texttt{L11-5v} and \texttt{L12-3v} probe settings achieved effective speckle suppression while preserving boundary definition, yielding sharp, well-defined edges, whereas images simulated under \texttt{C5-2v} appeared over-smoothed with softened edges and those simulated under \texttt{P4-2v} showed persistent granular artifacts and poor structural recovery.

These visual degradations are reflected quantitatively in Table~\ref{tab:transducer_results}. PSNR was highest on the in-distribution \texttt{L11-5v} probe, with \texttt{L12-3v} following closely. In contrast, \texttt{C5-2v} and \texttt{P4-2v} image quality degraded substantially: relative to \texttt{L11-5v}, PSNR dropped by $2.82$ dB on \texttt{C5-2v} and $3.87$ dB on \texttt{P4-2v}, with corresponding decreases in SSIM and increases in LPIPS. Notably, a $\sim$3 dB reduction in PSNR corresponds to roughly doubling the MSE, indicating that the poorer \texttt{C5-2v} and \texttt{P4-2v} outputs represent practically meaningful error increases rather than minor metric fluctuations.

\section{Discussion}
\label{sec:discussion}

\subsection{Performance on Real Ultrasound Images}
\label{sec:real_ultrasound_eval}

Because BrEaST Lesions USG is acquired under varied clinical conditions and is not tied to a specific imaging configuration, it provides a realistic test of model behavior under heterogeneous, in-the-wild ultrasound appearances. In the absence of paired ground truth, the qualitative comparisons in Figure~\ref{fig:referenceless_performance} suggest that IRSDE--Despeckle improves lesion conspicuity while suppressing speckle. In particular, our outputs better preserve lesion boundaries and layered tissue structure, whereas competing classical and learning-based approaches either leave residual speckle or introduce noticeable blurring that softens boundary definition.

The tendency to blur is apparent in generic natural-image restoration models, including denoisers such as DnCNN~\cite{DnCNN}, IRCNN~\cite{IRCNN}, and FFDNet~\cite{FFDNet}, as well as super-resolution-oriented networks such as IMDN~\cite{IMDN}, SRMD~\cite{SRMD}, and MSRResNet~\cite{MSRResNet}. A plausible interpretation of the blurring observed in these methods is that they reduce high-frequency noise but tend to either over-smooth fine anatomical structures or leave large speckle grains unresolved. This is likely because these models are designed and trained under assumptions suited for additive, approximately Gaussian noise, which does not match the behavior of multiplicative speckle. This causes them to suppress diagnostically relevant texture and flatten intensity transitions. Similarly, the ultrasound-specific baselines Speckle2Self~\cite{35_Li2025} and US Denoising GAN~\cite{gan_rw_zhang2022} produce overly homogenized outputs that erase tissue texture, limiting their ability to improve useful contrast.

Compared with these competing approaches, IRSDE--Despeckle leverages physically grounded training data with more realistic speckle statistics. As a result, our method better preserves background tissue structure while maintaining a cleaner lesion-to-background separation. This observation is, to a certain extent, consistent with the qualitative differences in CNR values overlaid in Figure~\ref{fig:referenceless_performance}. However, limitations remain. There are occasional hallucinated or exaggerated structures, particularly in regions of low original intensity or where sharp intensity transitions occur, such as when boundaries shift from well-defined to more diffuse. These artifacts suggest that although our model preserves details, additional regularization or data constraints may be needed to reduce the chance of generating structures that do not correspond to the true anatomy.

\subsection{Sensitivity to Simulation Parameters}

Our model was trained exclusively on simulated data generated using the \texttt{L11-5v} linear array transducer. While this fixes the acquisition process during training, it can bias the model toward probe-specific artifacts, potentially limiting generalizability. In particular, variations in array geometry and center frequency across transducer settings can shift speckle scale, resolution, and texture statistics beyond what is represented when training on a single simulation configuration.

Results in Section~\ref{subsec:cross_transducer_results} suggest that probe-dependent changes in the simulated acquisition contribute substantially to our model's generalization gap. The gap scales with the extent to which probe geometry and operating frequency deviate from the training configuration. \texttt{L11-5v} and \texttt{L12-3v} are both high-frequency linear arrays with similar frequency ranges and rectilinear beam geometry, which yields comparable point-spread functions and speckle statistics. In contrast, \texttt{C5-2v} and \texttt{P4-2v} operate at lower center frequencies and use curved or sector geometries, leading to larger speckle grains, different depth-dependent responses, and spatially varying resolution. These shifts change the appearance of the low-quality inputs and can propagate through the despeckling model, causing either over-smoothing or residual structured artifacts. Overall, these findings indicate that training on a single high-frequency linear probe provides limited coverage of the broader space of probe configurations. Multi-transducer training, frequency-aware augmentation, and domain adaptation strategies that explicitly account for probe geometry and frequency content would likely improve robustness across simulated and clinical scanners.

\begin{table}[t]
\centering
\caption{Despeckling performance of our despeckling model on simulated test images using the held out UMD dataset generated with different transducer settings. Metrics are reported as mean $\pm$ standard deviation over all test images.}
\label{tab:transducer_results}
\begin{tabular}{@{}lccc@{}}
\toprule
\textbf{Transducer Setting} & \textbf{PSNR$\uparrow$} & \textbf{SSIM$\uparrow$} & \textbf{LPIPS$\downarrow$} \\ 
\midrule
C5-2v   & $21.29 \pm 3.22$ & $0.47 \pm 0.11$ & $0.51 \pm 0.08$ \\
P4-2v   & $20.24 \pm 3.47$ & $0.43 \pm 0.11$ & $0.48 \pm 0.10$ \\
L12-3v  & $23.92 \pm 3.23$ & $0.58 \pm 0.12$ & $0.39 \pm 0.12$ \\
\texttt{L11-5v}  & $24.11 \pm 3.16$ & $0.56 \pm 0.10$ & $0.39 \pm 0.11$ \\
\bottomrule
\end{tabular}
\end{table}

\subsection{Toward Faster Diffusion-Based Despeckling}

A natural question is whether our diffusion-based despeckling model can achieve lower inference latency without compromising performance. Empirically, aggressive step reduction degrades reconstruction quality rapidly. Pushing step reduction further also makes training more difficult: In our experiment, when trained under such extreme step counts, optimization became unstable, the training loss plateaued at a significantly higher value, and the model failed to adequately learn the reverse process. This behavior is consistent with observations in the diffusion literature showing that overly coarse discretization weakens the expressivity of the learned transport map and reduces coverage of the continuous probability trajectory \cite{Song2020Score,Lu2022DPMSolver,Karras2022EDM}.

One promising direction for improving the efficiency of diffusion-based despeckling models is to accelerate both training and inference through alternative formulations of the generative process. On the training side, alternative parameterizations such as $v$ prediction, where the model predicts a velocity target rather than raw noise, can improve numerical conditioning across noise levels and lead to more stable and faster convergence \cite{Salimans2022ProgressiveDistillation,Karras2022EDM}. Alternatively, diffusion sampling can be cast as a deterministic probability flow ODE, removing inference-time noise and enabling more efficient integration \cite{Song2020Score}. Inference speed can also potentially be increased without retraining by adopting more efficient sampling schemes, for example, Denoising Diffusion Implicit Models \cite{Song2020DDIM}. Rectified flow, in addition, can be utilized to simplify the reverse diffusion path by encouraging straight transport trajectories between HQ and LQ samples \cite{Liu2022FlowStraightFast}.


\begin{table}[t]
\centering
\caption{Quantitative despeckling performance and average inference time per $128\times128$ input image for models trained with different numbers of reverse diffusion steps $T$. Metrics are reported on the held-out UMD test set.}
\label{tab:timesteps_results}
\begin{tabular}{@{}lcccc@{}}
\toprule
\textbf{Sampling Steps $T$} & \textbf{PSNR$\uparrow$} & \textbf{SSIM$\uparrow$} & \textbf{LPIPS$\downarrow$} & \textbf{Time (s)$\downarrow$} \\ 
\midrule
$100$ & 20.95 & 0.60 & 0.30 & 1.04 \\
$50$ & 20.49 & 0.58 & 0.30 & 0.52 \\
$20$ & 17.90 & 0.42 & 0.43 & 0.21 \\
$10$ & 14.17 & 0.22 & 0.66 & 0.11 \\
\bottomrule
\end{tabular}
\end{table}

\section{Conclusions}

In this work, we introduced a novel ultrasound despeckling method based on the IRSDE framework. We leveraged MRI scans as high-quality, speckle-free targets and simulated corresponding low-quality ultrasound images that follow realistic ultrasound physics using the MUST simulation toolbox. Using this synthesized large ultrasound dataset, our model learned to remove speckle corresponding to ultrasound physics and restore MRI-like structural detail, outperforming state-of-the-art despeckling methods in both quantitative metrics and qualitative assessment. Our uncertainty analysis demonstrated that the model’s prediction uncertainty closely tracks reconstruction error, suggesting that ensemble-based uncertainty estimates could support downstream clinical interpretation. Future work will focus on accelerating inference for real-time deployment, expanding training to enhance generalization, and extending the framework to ultrasound video with temporally consistent structure restoration.

\section*{Acknowledgements}
This work was supported by the National Institutes of Health (grants R01DE033175 and R56DE033175).




\section*{Data availability}
The data and code supporting the findings of this study will be made publicly available upon publication.

\bibliographystyle{elsarticle-harv}
\bibliography{refs}   

\clearpage
\onecolumn
\appendix
\setcounter{equation}{0}
\setcounter{figure}{0}
\setcounter{table}{0}
\section{Additional Experimental Results}
\label{app:additional_experimental_results}

\subsection{Visualizations of Experimental Results with Ground-truth}
\label{app:gt_results}

\begin{samepage}
We report additional qualitative results on the held-out UMD simulated ultrasound test set \cite{UMD2024}. As described in Section~\ref{sec:results_discussion}, LQ images were simulated from HQ targets using MUST with the \texttt{L11-5v} transducer setting \cite{MUST}. The test set contains 400 paired HQ--LQ samples, which are consisted of 200 samples simulated from full 2D MRI slices, and 200 samples from cropped patches of the corresponding full 2D views.

\begin{figure}[H]
\centering
\includegraphics[width=0.8\linewidth]{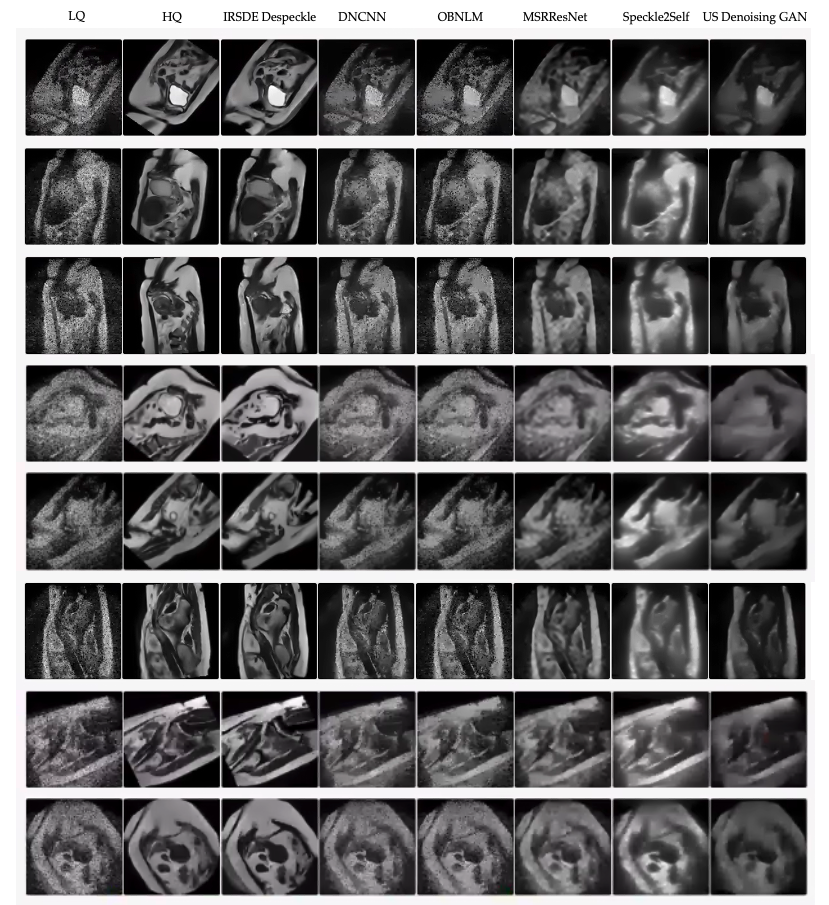}
\caption{Qualitative comparison on ultrasound data simluated from full-view MRI 2D slices. Columns show the low-quality (LQ) speckled input, high-quality (HQ) reference patch, and outputs from baseline methods and IRSDE Despeckle. Due to space limitations, we show selected learning-based baselines. The examples illustrate tradeoffs between speckle suppression and structural preservation.}
\label{fig:app_whole_gt}
\end{figure}
\end{samepage}

\begin{figure}[p]
\centering
\includegraphics[width=0.8\linewidth]{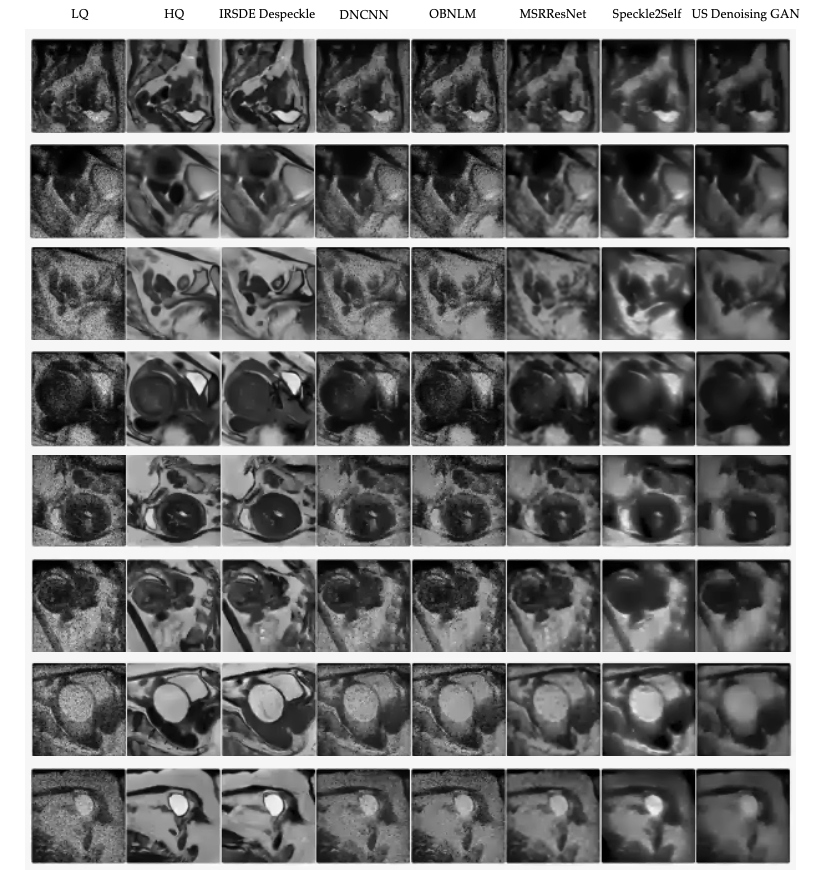}
\caption{Qualitative comparison on ultrasound data simulated from ROI patches of MRI 2D slices. Columns show the low-quality (LQ) speckled input, high-quality (HQ) reference patch, and outputs from baseline methods and IRSDE Despeckle. Due to space limitations, we show selected learning-based baselines. The examples illustrate tradeoffs between speckle suppression and structural preservation.}
\label{fig:app_patch_gt}
\end{figure}

\clearpage

\subsection{Experimental Results with Referentless Quality Metrics}
\label{app:referenceless_metrics}

We considered four \emph{reference-less} image quality metrics that can be computed on real ultrasound images without paired ground truth. These metrics were used to characterize desirable aspects of despeckling performance, namely tumor-to-background contrast, background homogeneity, and perceptual quality.

Contrast-to-noise ratio (CNR) quantifies how strongly a foreground region (e.g., a tumor) stands out from its surrounding background. Given a foreground region of interest (ROI) and a background ROI, we computed
\begin{equation}
\label{eq:app_cnr}
\mathrm{CNR}
= \frac{\lvert\mu_{\mathrm{fg}}-\mu_{\mathrm{bg}}\rvert}{\sqrt{\sigma_{\mathrm{fg}}^{2}+\sigma_{\mathrm{bg}}^{2}+\epsilon}},
\end{equation}
where $\mu_{\mathrm{fg}}$ and $\sigma_{\mathrm{fg}}$ denote the mean and standard deviation within the foreground ROI, $\mu_{\mathrm{bg}}$ and $\sigma_{\mathrm{bg}}$ are the corresponding statistics in the background ROI, and $\epsilon$ is a small constant for numerical stability. This formulation is consistent with classical tumor detectability measures \cite{RodriguezMolares2019gCNR}.

When available, the foreground ROI was given by the tumor segmentation mask provided by BrEaST tumors USG dataset. We selected the background ROI from a depth-matched band spanning the tumor depth. Within this band (excluding the tumor), we tiled the remaining region into non-overlapping patches. For each patch, we computed intensity statistics and Sobel gradient energy, and retained only locally homogeneou. We then pooled the five most homogeneous patches into a composite background ROI. Figure~\ref{fig:referenceless_performance} shows an example of the foreground/background ROI selection.

Equivalent number of looks (ENL) measures speckle homogeneity in a nominally uniform background region. It is defined as \cite{Baselice2017StatisticalSimilarityDespeckling}
\begin{equation}
\label{eq:app_enl}
\mathrm{ENL} = \frac{\mu_{\mathrm{bg}}^{2}}{\sigma_{\mathrm{bg}}^{2}+\epsilon},
\end{equation}
with higher ENL indicating reduced relative variance and smoother appearance. Ignoring $\epsilon$, ENL is the squared coefficient of variation inverted, i.e., $\mathrm{ENL} \approx (\mu_{\mathrm{bg}}/\sigma_{\mathrm{bg}})^2$.

Speckle signal-to-noise ratio (Speckle SNR) is closely related to ENL under classical speckle statistics \cite{Wagner1983}, is expressed in decibels as:
\begin{equation}
\label{eq:app_speckle_snr}
\mathrm{SNR}_{\mathrm{dB}} = 20\log_{10}\Bigl(\frac{\mu_{\mathrm{bg}}}{\sigma_{\mathrm{bg}}+\epsilon}\Bigr).
\end{equation}

We used $20\log_{10}(\cdot)$ because the ratio inside the logarithm is an amplitude-like quantity.

To capture generic perceptual quality, we also computed BRISQUE, a no-reference perceptual quality metric based on deviations from natural image statistics \cite{Mittal2012BRISQUE}. BRISQUE metric was originally developed for natural images and its absolute values should therefore be interpreted cautiously in the ultrasound context.

One limitation of referenceless metric evaluation is that, although CNR, ENL, and Speckle SNR are widely used to quantify contrast and speckle suppression, they require cautious interpretation \cite{48_Narayanan2009}. CNR is highly dependent on ROI selection and is sensitive to both foreground boundary accuracy and background placement. As a result, changes in CNR may reflect intensity scaling effects rather than genuine improvements in tissue structure visibility. Similarly, higher ENL or Speckle SNR values do not necessarily indicate clinically meaningful improvement. Aggressive smoothing can artificially increase background homogeneity, inflating ENL and Speckle SNR scores while simultaneously blurring diagnostically relevant structures and reducing tumor contrast. These metrics do not explicitly penalize the loss of fine features or structural detail. Perceptual metrics such as BRISQUE were developed for natural images \cite{Mittal2012BRISQUE} and are not tailored to ultrasound image statistics and thus may not correlate with image quality when applied to speckle-reduced ultrasound images. 

\begin{table}[H]
\centering
\caption{Referenceless ultrasound image quality metrics (mean $\pm$ std) across all images. Best value in each column is in bold.}
\label{tab:referenceless_metrics}
\begin{tabular}{@{}lcccc@{}}
\toprule
\textbf{Method} & \textbf{CNR$\uparrow$} & \textbf{ENL$\uparrow$} & \textbf{Speckle SNR (dB)$\uparrow$} & \textbf{BRISQUE$\downarrow$} \\
\midrule
DnCNN                    & 0.80 $\pm$ 0.48 & 8.68 $\pm$ 6.19  & 8.25 $\pm$ 3.45  & 38.28 $\pm$ 11.79 \\
FFDNet                   & 0.74 $\pm$ 0.44 & 7.08 $\pm$ 4.56  & 7.44 $\pm$ 3.42  & 28.54 $\pm$ 8.73  \\
IMDN                     & 0.85 $\pm$ 0.49 & \textbf{10.08 $\pm$ 7.07} & \textbf{8.76 $\pm$ 3.81}  & 58.16 $\pm$ 8.70  \\
IRCNN                    & 0.76 $\pm$ 0.47 & 9.14 $\pm$ 6.61  & 8.40 $\pm$ 3.60  & 27.61 $\pm$ 7.21  \\
MSRResNet                & 0.85 $\pm$ 0.49 & 9.96 $\pm$ 7.01  & 8.71 $\pm$ 3.81  & 56.83 $\pm$ 7.42  \\
SRMD                     & 0.85 $\pm$ 0.49 & 9.99 $\pm$ 7.01  & 8.72 $\pm$ 3.82  & 60.20 $\pm$ 10.72 \\
Speckle2Self             & \textbf{0.88 $\pm$ 0.51} & 5.13 $\pm$ 3.32  & 6.23 $\pm$ 2.95  & 53.37 $\pm$ 8.01  \\
Ultrasound Denoising GAN & 0.82 $\pm$ 0.48 & 6.57 $\pm$ 5.42  & 7.09 $\pm$ 3.16  & 37.81 $\pm$ 7.93  \\
IRSDE Despeckle          & 0.54 $\pm$ 0.32 & 3.63 $\pm$ 2.10  & 4.79 $\pm$ 2.83  & \textbf{27.11 $\pm$ 10.85} \\
\bottomrule
\end{tabular}
\end{table}

\begin{figure}[p]
\centering
\setlength{\tabcolsep}{0pt}
\renewcommand{\arraystretch}{0}
\begin{tabular}{@{}c@{}}
\includegraphics[width=0.95\linewidth]{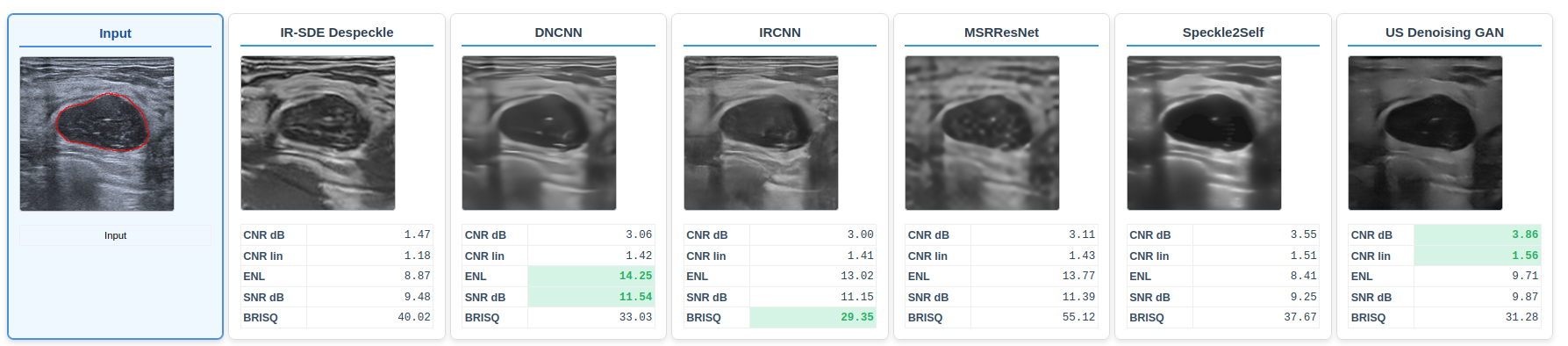}\\[-2pt]
\includegraphics[width=0.95\linewidth]{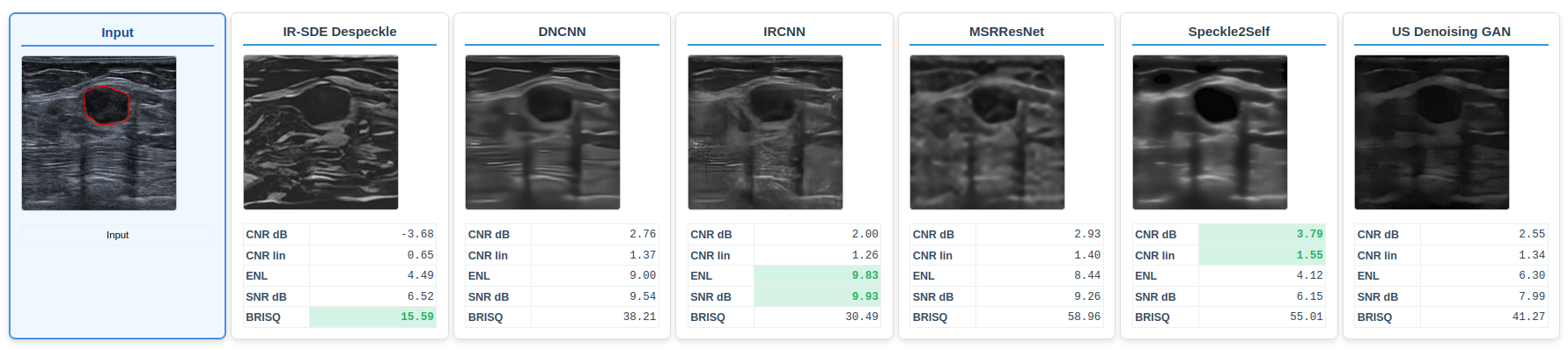}\\[-2pt]
\includegraphics[width=0.95\linewidth]{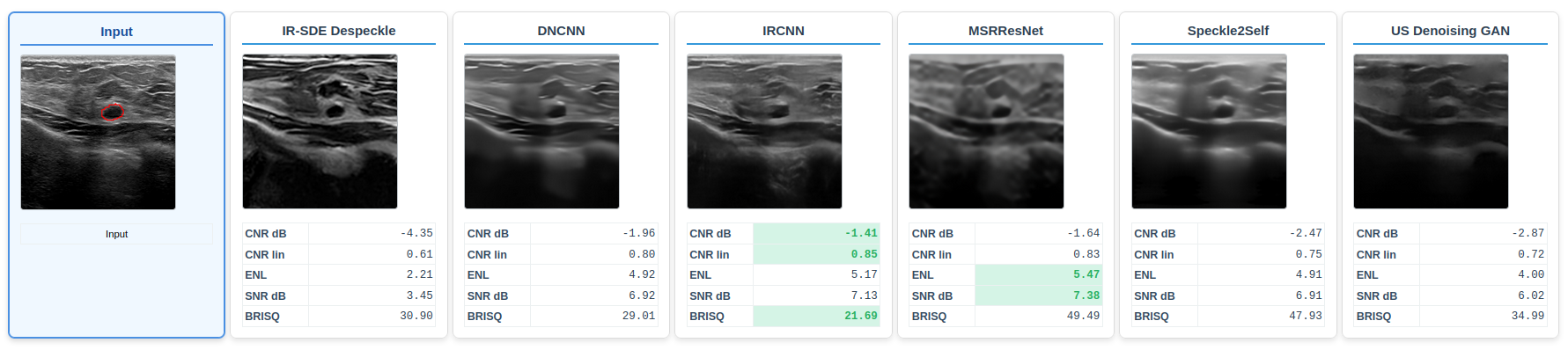}\\[-2pt]
\includegraphics[width=0.95\linewidth]{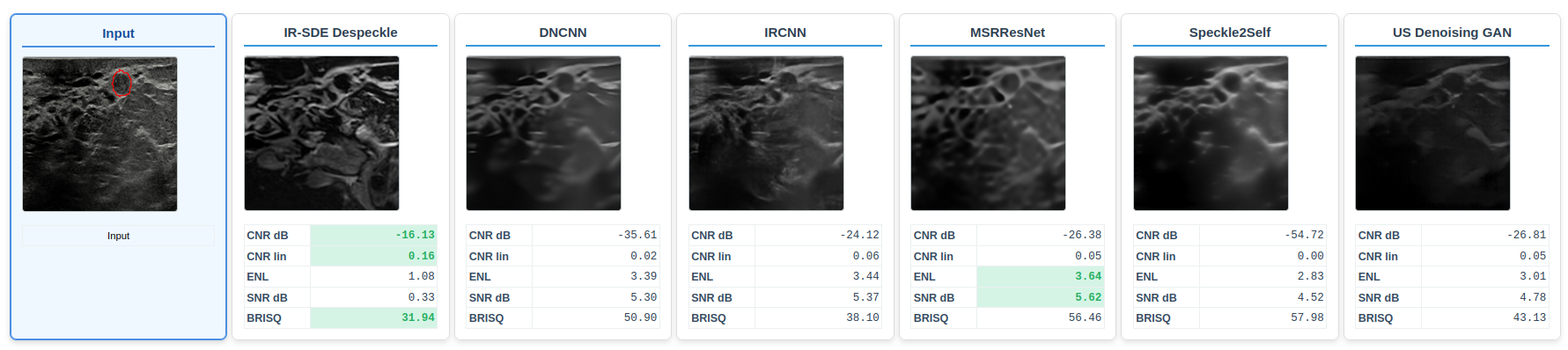}\\[-2pt]
\includegraphics[width=0.95\linewidth]{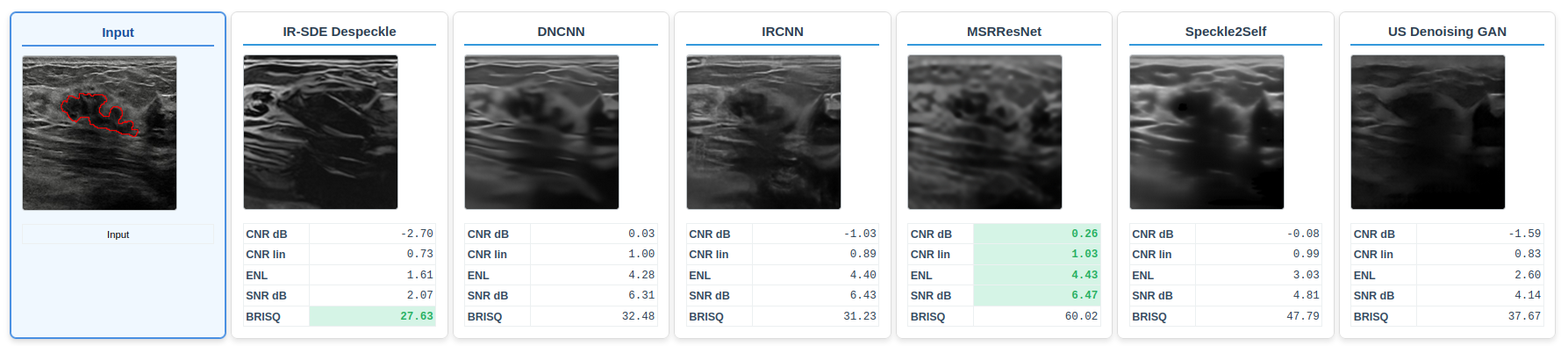}
\end{tabular}
\caption{Additional referenceless qualitative examples used for image quality assessment on the BrEaST tumors USG dataset. Each row corresponds to a separate test case, showing the input image and the outputs of the compared despeckling methods, along with the associated referenceless quality metrics when available. The red contour in the input image indicates the tumor annotation provided by the dataset. For clarity, we omit the background ROI overlay in this figure; an example showing both the foreground and background ROIs is provided in the main text (Fig.~\ref{fig:referenceless_performance}). Quantitative metrics include CNR, ENL, Speckle SNR, and BRISQUE, with green highlights marking the best value per metric across methods for that case. While some methods achieve higher scores on individual metrics, visual inspection reveals notable differences in structure preservation, edge fidelity, and tumor appearance, underscoring that metric improvements do not always align with perceptual or clinically relevant quality. These examples motivate interpreting referenceless metrics in conjunction with case-by-case visualization.}
\label{fig:app_referenceless_cases}
\end{figure}

Table~\ref{tab:referenceless_metrics} highlights the limitations and tradeoffs inherent in the aformenteiond referenceless metrics evaluation. Methods such as IMDN, and SRMD achieve the highest ENL and Speckle SNR values due to aggressive smoothing. Such gains may not necessarily translate to improved tumor visibility or structural preservation, since these metrics do not penalize loss of fine anatomical detail. IRSDE Despeckle exhibits lower ENL and Speckle SNR scores compared, and this is consistent with a more conservative despeckling behavior. IRSDE Despeckl does achieve the best BRISQUE score overall. We hypothesize that this improvement is largely driven by the MRI-like target appearance encouraged during training, which produces cleaner, more globally consistent textures that align with natural image statistics. Given these considerations, we treat these numerical results as only indications of despeckling capability rather than definitive measures of quality, and place greater emphasis on qualitative, case-by-case visualization in Fig.~\ref{fig:app_referenceless_cases} below to assess feature preservation.

\FloatBarrier

\subsection{Utilized Datasets}
\label{app:datasets}

To encourage generalization, we used multiple publicly available datasets spanning diverse anatomies. We constructed the MUST-simuated training set from a wide range of MRI datsets including MRNet Knee MRIs \cite{MRNet}, Duke Liver Dataset v2 \cite{DukeLiverV2}, BrainMetShare 3 \cite{BrainMetShare}, and Duke Breast Cancer MRI \cite{DukeBreastCancerMRI}, and we tested out of distribution performance on the Uterine Myoma MRI Dataset (UMD), which sampled distinct pelvic anatomy and was processed with the same pipeline as the training set. 

We additionally evaluated on real ultrasound images using BrEaST Lesions USG \cite{breastlesions_usg}, a breast ultrasound benchmark that served as an in the wild robustness test under clinically heterogeneous, operator dependent imaging conditions.

\paragraph{\textbf{MRNet Knee MRIs}}
MRNet comprises 1,370 clinical knee MRI examinations. Each examination contains three series acquired in the sagittal, coronal, and axial planes and includes study level labels for abnormality, ACL tear, and meniscal tear derived from clinical reports. The knee anatomy includes strong intensity gradients and crisp boundaries among bone and cartilage, which complements softer tissue dominant MRI datasets and broadens the appearance distribution seen during training.

\paragraph{\textbf{Duke Liver Dataset v2}}
Duke Liver Dataset v2 provides 2,146 abdominal MRI series from 105 patients, with a substantial fraction exhibiting cirrhotic features. A subset of 310 series includes expert liver segmentation masks; we do not use these masks for supervision. Nonetheless, the dataset contributes diverse abdominal anatomy and heterogeneous routine liver MRI appearances.

\paragraph{\textbf{BrainMetShare-3}}
BrainMetShare-3 includes 156 whole brain MRI studies from patients with brain metastases and provides multi modal pre contrast and post contrast sequences. it contains lesion bearing studies and multi sequence brain MRI, which increases diversity in fine scale texture, contrast patterns, and pathology related appearances.

\paragraph{\textbf{Duke Breast Cancer MRI}}
Duke Breast Cancer MRI consists of 922 patient studies with invasive breast cancer and pre operative dynamic contrast enhanced breast MRI. The images are predominantly soft tissue and therefore complements higher contrast anatomy in other datasets by adding soft tissue dominant appearance variability that is common in clinical applications.

\paragraph{\textbf{Uterine Myoma MRI Dataset}}
Uterine Myoma MRI Dataset (UMD) contains sagittal T2 weighted pelvic MRI from 300 patients with uterine myoma, with pixel level labels for uterine wall, uterine cavity, myoma, and nabothian cyst. We use UMD only for evaluation and treat it as anatomically out of distribution relative to the training set. For a controlled comparison, we process UMD with the same pipeline used for training data to generate simulated ground truth targets.

\paragraph{\textbf{BrEaST Lesions USG}}
BrEaST Lesions USG includes 256 breast ultrasound scans, including benign and malignant lesions and a small number of normal cases. The datasets offers clinically heterogeneous and operator dependent ultrasound appearance, including speckle, shadowing, and device setting variability. It also includes pixel level tumor masks with accompanying metadata, and we use the provided masks to define the foreground region of interest.

\end{document}